%% file: draft.tex
\documentclass[10pt, a4paper]{article}

\usepackage{comment}
\usepackage{siunitx}
\usepackage{booktabs}
\usepackage{multirow}
\usepackage{enumitem}
\usepackage{tikz}
\usetikzlibrary{shapes.geometric, arrows.meta}
\usepackage{appendix}
\usepackage{tcolorbox}
\usepackage{xcolor}
\usepackage{caption} 
\usepackage{subcaption}
\usepackage{colortbl}
\definecolor{lightestgray}{gray}{0.95}

\usepackage{placeins}
\usepackage{textcomp}  
\usepackage{scalerel}  
\usepackage{relsize}
\def\affiluhel{\scalerel*{\includegraphics[scale=0.4]{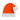}}{\textrm{\footnotesize\textbigcircle}}}
\def\affileqcontrib{\hspace{0.1em}\raisebox{0.1ex}{\scalerel*{\includegraphics[scale=0.4]{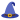}}{\textrm{\footnotesize\textbigcircle}}}}
\def\affillorraine{\scalerel*{\includegraphics[scale=0.4]{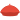}}{\textrm{\footnotesize\textbigcircle}}}
\def\affilcharles{\scalerel*{\includegraphics[scale=0.4]{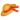}}{\textrm{\footnotesize\textbigcircle}}}
\def\affilindependent{\scalerel*{\includegraphics[scale=0.4]{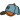}}{\textrm{\footnotesize\textbigcircle}}}
\def\affilsorbparis{\scalerel*{\includegraphics[scale=0.4]{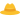}}{\textrm{\footnotesize\textbigcircle}}}
\def\affiliiit{\scalerel*{\includegraphics[scale=0.4]{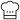}}{\textrm{\footnotesize\textbigcircle}}}
\def\affiltorino{\scalerel*{\includegraphics[scale=0.4]{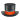}}{\textrm{\footnotesize\textbigcircle}}}
\def\affilubc{\scalerel*{\includegraphics[scale=0.4]{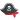}}{\textrm{\footnotesize\textbigcircle}}}
\def\affilchosun{\scalerel*{\includegraphics[scale=0.4]{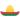}}{\textrm{\footnotesize\textbigcircle}}}
\def\affiltromso{\scalerel*{\includegraphics[scale=0.4]{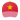}}{\textrm{\footnotesize\textbigcircle}}}

\usepackage[final]{lrec2026} 

\usepackage{cleveref}

\title{Confabulations from ACL Publications (CAP): \\ A Dataset for Scientific Hallucination Detection}

\name{
\textbf{Federica Gamba \textsuperscript{\hspace{-0.5em}\affilcharles\affileqcontrib}} \qquad
  \textbf{Aman Sinha\textsuperscript{\affillorraine\affileqcontrib}} \qquad
  \textbf{Timothee Mickus\textsuperscript{\affiluhel}}
\\
    \textbf{Ra\'ul V\'azquez\textsuperscript{\affiluhel}} \qquad
    \textbf{Patanjali Bhamidipati\textsuperscript{\affiliiit}} \qquad
    \textbf{Claudio Savelli\textsuperscript{\affiltorino}} 
\\     
    \textbf{Ahana Chattopadhyay\textsuperscript{\affillorraine}} \qquad
    \textbf{Laura Zanella\textsuperscript{\affilindependent}} \qquad
    \textbf{Yash Kankanampati \textsuperscript{\hspace{-0.5em}\affilsorbparis}} 
\\
    \textbf{Binesh Arakkal Remesh \textsuperscript{\hspace{-0.5em}\affillorraine}} \qquad
   \textbf{Aryan Chandramania\textsuperscript{\hspace{-0.2em}\affiliiit}} \qquad
  \textbf{Rohit Agarwal \textsuperscript{\hspace{-0.5em}\affiltromso}}
\\ 
  \textbf{Chuyuan Li \textsuperscript{\hspace{-0.5em}\affilubc}} \qquad
    \textbf{Ioana Buhnila \textsuperscript{\hspace{-0.5em}\affilchosun}} \qquad
    \textbf{Radhika Mamidi \textsuperscript{\hspace{-0.5em}\affiliiit}}\\
 \textsuperscript{\affileqcontrib}{\small These authors have equal contribution. }
}
\address{
    \textsuperscript{\affillorraine\hspace{0.1em}}Universit\'e  de Lorraine, France; \quad
    \textsuperscript{\affilcharles\hspace{0.1em}}Charles University, Prague; \quad
    \textsuperscript{\affilindependent\hspace{0.1em}}Independent Researcher;
\\
    \textsuperscript{\affiluhel\hspace{0.1em}}University of Helsinki, Finland; \qquad
    \textsuperscript{\affilsorbparis\hspace{0.1em}}Universit\'e Sorbonne Paris Nord, France; 
\\
    \textsuperscript{\affiliiit\hspace{0.1em}}IIIT Hyderabad, India; \quad
    \textsuperscript{\affiltorino\hspace{0.1em}}Politecnico di Torino, Italy; \quad
    \textsuperscript{\affiltromso\hspace{0.1em}}UiT Tromso, Norway;
    \\ 
    \textsuperscript{\affilubc\hspace{0.1em}}University of British Columbia, Vancouver, Canada \qquad
    \textsuperscript{\affilchosun\hspace{0.1em}}Chosun University, South Korea
\\ [0.5em]
  \small{
    \textbf{Correspondence:} \{\href{mailto:gamba@ufal.mff.cuni.cz}{gamba@ufal.mff.cuni.cz}, \href{mailto:aman.sinha@univ-lorraine.fr}{aman.sinha@univ-lorraine.fr}\}
  }
}

\newcommand\orgname[1]{\noindent\textbf{#1}} 
\abstract{
We introduce the CAP (Confabulations from ACL Publications) dataset, a multilingual resource for studying hallucinations in large language models (LLMs) within scientific text generation. CAP focuses on the scientific domain, where hallucinations can distort factual knowledge, as they frequently do. In this domain, however, the presence of specialized terminology, statistical reasoning, and context-dependent interpretations further exacerbates these distortions, particularly given LLMs’ lack of true comprehension, limited contextual understanding, and bias toward surface-level generalization. CAP operates in a cross-lingual setting covering five high-resource languages (English, French, Hindi, Italian, and Spanish) and four low-resource languages (Bengali, Gujarati, Malayalam, and Telugu). The dataset comprises 900 curated scientific questions and over 7,000 LLM-generated answers from 16 publicly available models, provided as question–answer pairs along with token sequences and corresponding logits. Each instance is annotated with a binary label indicating the presence of a scientific hallucination, denoted as a factuality error, and 
a fluency label, capturing issues in the linguistic quality or naturalness of the text. 
CAP is publicly released to facilitate advanced research on hallucination detection, multilingual evaluation of LLMs, and the development of more reliable scientific NLP systems.
\\ \newline  }

\begin{document}

\maketitleabstract

\section{Introduction}
As the prevalence of large language model (LLM) technology grows, so do concerns about its reliability and trustworthiness.
This state of affairs stems from the general ambivalence of these systems when it comes to the truthfulness of their outputs \citep{Hicks2024, deemter-2024-pitfalls, perez-etal-2023-discovering} --- these models grow more fluent, but they need not output sentences that are factually correct, which gives rise to fluent but factually false outputs, or `hallucinations'.
Remarkably, researchers interested in hallucinations often emphasize hallucinations as issues of factuality: \citet{kalai2025languagemodelshallucinate} define hallucinations as ``plausible falsehoods'', whereas \citet{deemter-2024-pitfalls} provides a taxonomy based on logical contradictions.
Such approaches commonly assume that LLMs consistently produce fluent and coherent text across languages. This assumption, however, does not necessarily hold in the context of \textit{low-resource languages}.

In languages with limited representation in training corpora, LLMs frequently exhibit reduced fluency, degraded coherence, and a higher incidence of semantically or syntactically flawed outputs \citep{dargis-etal-2024-evaluating}. These limitations can result in hallucinations that go beyond factual inaccuracy, including structural errors, semantic drift, or outright nonsensical generations. The scarcity of high-quality datasets for such languages constrains both model training and evaluation, making it difficult to assess and ensure reliability.
%
Together, these observations underscore the need for language technologies that account for linguistic diversity and for benchmark resources that can capture model behavior more accurately across underrepresented languages.

\begin{figure}
    \centering
    \includegraphics[width=\columnwidth, trim= 1em 1em 1em 0, clip]{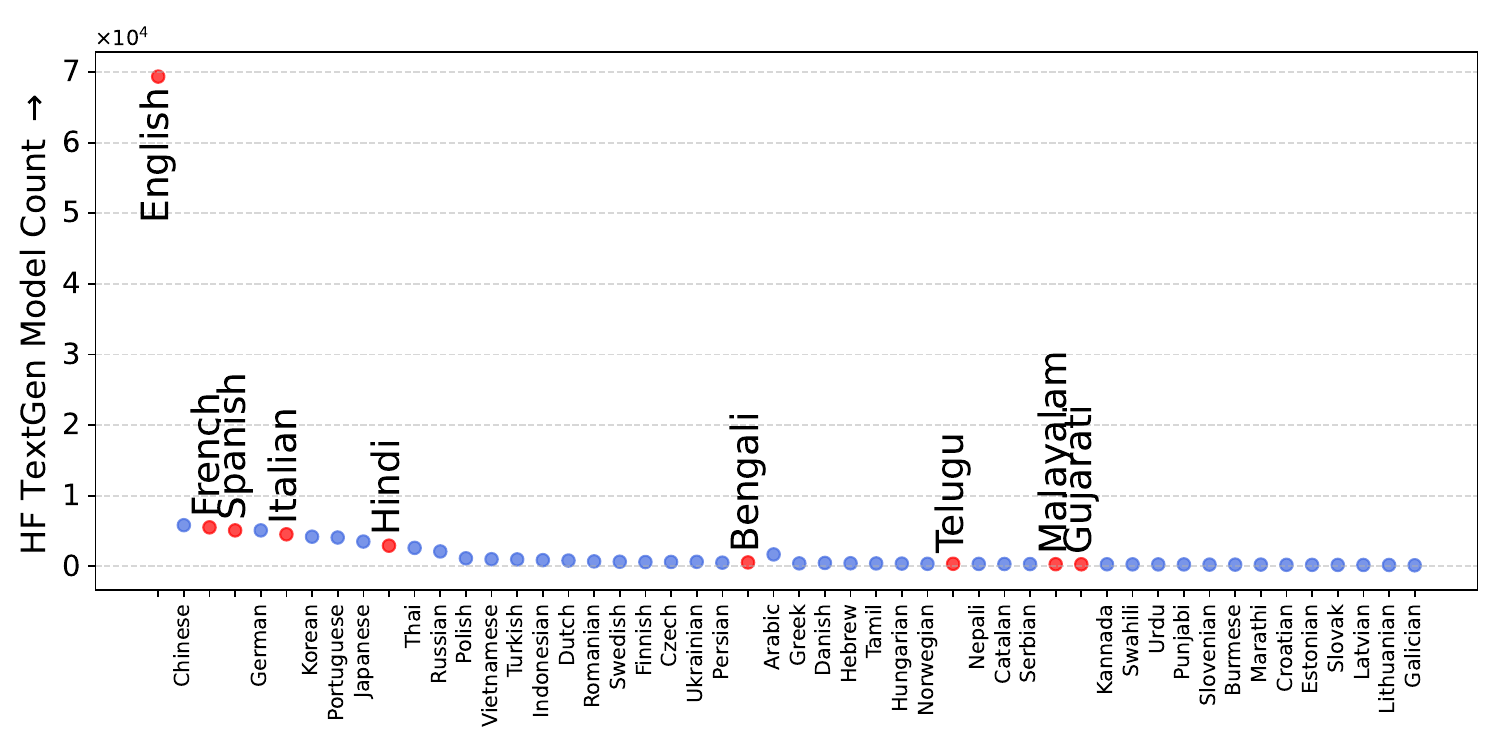}
    \caption{Distribution of publicly available text generation models on HuggingFace (HF) website as of October 22nd, 2025.}
    \label{fig:stats-hf-models}
\end{figure}

We argue that the mechanisms underlying hallucinatory behaviors in LLMs thus stand at the crossroads of the challenges in fluency and factuality. Consequently, we can expect that hallucinations will look different for languages for which models tend to be less fluent \citep{vazquez-etal-2025-semeval} or domains that require more specialized knowledge \citep{george-stuhlmueller-2023-factored}.  Indicators of disparitties in access to NLP technology, can be illustrated by the variety of tools available for a given language. As we show in \Cref{fig:stats-hf-models}, the vast majority of languages are undeserved causing a clear imbalance between e.g.\ French and Telugu or Bengali.

In this paper, we introduce \textbf{CAP}, a multilingual question-answering dataset designed to evaluate hallucinations along both fluency and factuality dimensions across nine languages, using scientific publications retrieved from the ACL Anthology. 
CAP comprises a total of 100 unique questions per language, with eight LLM-generated response annotations per question (\Cref{tab:dataset-stats}). 
We conduct extensive experiments to evaluate the performance of six representative hallucination detection baselines spanning reference-based and reference-free paradigms. Our results indicate that existing hallucination detection tools are generally ill-suited for CAP, with most configurations performing near or below random. In addition to benchmarking model performance, we leverage CAP’s fine-grained annotation layers to investigate the root causes of hallucinations: in particular, we review the effects of citation counts (as a proxy for notoriety), question type, and context complexity.  Our analyses show that linguistic cues such as the type of questions asked or the context used as input for hallucination detection tools tend to explain hallucination rates better than citation counts.

Overall, the contributions of this paper include: 
\begin{itemize}
    \item \textbf{CAP (Confabulations from ACL Publications)}, a novel scientific hallucination dataset comprising question-answer pairs for nine languages, including five high-resource and four low-resource languages; and
    \item \textbf{A benchmark evaluation of existing hallucination detection tools} with respect to our proposed definition of hallucination phenomena, showcasing the remaining challenges and suggesting directions for future research in hallucination detection.
\end{itemize}

\section{Related Works}

LLMs often generate hallucinations: outputs that appear coherent and well-formed but contain factual inaccuracies. As discussed in several recent surveys \cite{ji-etal-2023-survey, huang-etal-2024-survey}, this issue raises concerns about the reliability of LLMs in knowledge-intensive domains such as scientific writing. In particular, \citet{george-stuhlmueller-2023-factored} highlights how LLMs can generate unsupported claims, invented references, and factually incorrect statements when summarizing or rewriting scholarly articles.

Recent work has focused on factuality evaluation, aiming to assess how well model outputs align with reference information. Early benchmarks \cite{kryscinski2019evaluating, wang2020asking} assessed factual consistency in English summarization using entailment or QA-based proxies. Subsequent studies \cite{wadden-etal-2022-scifact, wadden-etal-2022-multivers} showed that domain-specific factuality, particularly in scientific writing, cannot be reduced to textual similarity but requires grounding in specialized terminology and evidence retrieval. \citet{qi2023cross} analyzed the cross-lingual consistency of factual knowledge in multilingual LLMs, finding that many models achieve low factual alignment across languages and rely heavily on lexical overlap rather than language-independent representations. 
Together, these findings reveal that current LLMs struggle to maintain factual coherence across scientific domains and languages.

Several benchmarks have been proposed to evaluate hallucination in LLMs, such as \citet{mickus-etal-2024-semeval} for general NLG tasks and \citet{yasunaga2019scisummnet, wadden-etal-2022-scifact} for scientific summarization and claim verification. However, all these resources are limited to English. \citet{vazquez-etal-2025-semeval} denotes a recent effort toward evaluating hallucination in multilingual environments, reflecting the increasing attention to extend factuality assessment to less represented languages. Building on \citeauthor{vazquez-etal-2025-semeval}'s work, \citet{rykov2025modelslielearnmultilingual} propose a large-scale and multilingual dataset of automatically annotated annotations and demonstrate that such a resource can bolster the efficacy of hallucination detection tools.

\section{CAP Dataset}

\subsection{Overview}


The CAP dataset is a multilingual resource designed to evaluate hallucinations in LLM outputs for scientific text generation. It is comprised of nine languages in total: five high-resource languages, including English (en), French (fr), Hindi (hi), Italian (it), and Spanish (es); and four low-resource Indic languages, including Bengali (bn), Gujarati (gu), Malayalam (ml), and Telugu (te).

\input{Figures/sample-shroomcap-fix}
\Cref{fig:example} presents an example instance from CAP Malayalam data split. Each instance in the dataset corresponds to a question-answer pair associated with a scientific natural language processing (NLP) paper. These are provided with a comprehensive set of metadata fields, when available, including an index identifier, paper title, abstract, DOI, and URL, as well as all the authors' information, including given name and surname recorded separately to facilitate prompt construction. In addition to the metadata about the question, we also provide metadata about the generated answer (referred to as \texttt{output\_text} in \Cref{fig:example}) from the LLM and the LLM model. This includes a tokenized version of the generated answer, the associated logits, and, for the LLM model, we provide the model identifier, model configuration, and the prompt used. Each instance contains two labels, namely, \texttt{has\_fluency\_mistakes} (corresponding to fluency) and \texttt{has\_factual\_mistakes} (corresponding to hallucination) for the generated answer. Altogether, this information enables detailed investigations of model behavior across languages and scientific domains.



\subsection{CAP Creation Workflow}
\label{sec:workfolw}

The data creation process comprises three main stages. First, we curate a cohort of scientific papers. Next, human annotators manually compose questions based on these papers, which are subsequently presented to LLMs to generate corresponding answers.

\paragraph{Scientific Paper Collection: }
We began by \emph{collecting papers} from the ACL Anthology, compiling a set of 293 award-winning papers in NLP from 1995 to 2024. 
These papers are more likely to be cited and discussed, making them a strong source for hallucination detection.
For each selected paper, we extracted available metadata, including the title, abstract, DOI, URL, and list of authors. This structured metadata provides a comprehensive representation of each paper and serves as the foundation for subsequent question creation and answer generation steps.
    
\paragraph{Question Creation:}
For each language, annotators were provided with a randomly selected subset of $100$ papers (in English) and were instructed to manually create one question for each paper in their assigned language.
Since the sampling was performed independently for each language, the resulting subsets differ, meaning that the list of papers annotated in one language does not necessarily overlap with those used in another.
For any paper appearing in multiple languages, questions are not simple translations; rather, they are independently crafted by annotators in each language to reflect language-specific perspectives and nuances.
Annotators were supported by a script to record each question, providing the corresponding paper link.


\paragraph{Response Generation using LLMs:}
For each question, we generated multiple responses (ranging between 6 to 18) using publicly available, instruction-tuned LLMs capable of handling the target languages (see details in Section~\ref{subsec:stats}).
Multiple outputs per question were obtained by varying generation hyperparameters, including top-$p$, top-$k$, and temperature. Specifically, we use: 
(1) default setting; (2) top-$p$=0.9, top-$k$=50, temperature=0.1; (3) top-$p$=0.95, top-$k$=50, temperature=0.2.
We also vary the input context provided in the prompt used for answer generation by optionally adding abstract as an additional information.


\subsection{Annotation Details}
\label{subsec:annotation}

\paragraph{Human Annotators:}
Question creation and LLM-generated answer annotations were carried out by a team of ten annotators. 
All of them are native speakers of the assigned language and possess strong NLP backgrounds, ranging from graduate students to postdoctoral researcher,
ensuring domain expertise in the annotation process.
The number of annotators across languages is as follows: English (3), French (2), Hindi (2), Italian (1), Spanish (2), Bengali (1), Gujarati (1), Malayalam (1), and Telugu (1).
Annotators were instructed to formulate a question using any combination of the title, abstract, full paper, or author information, based on their preference.

\begin{figure}[ht!]
    \centering
\begin{tikzpicture}
\node[anchor=north west] (box) at (0, 0) {
    \begin{tcolorbox}[colframe=white!75, colback=gray!10, width=0.95\columnwidth]
    In the article titled [\texttt{title}] by  [\texttt{last}],  [\texttt{first}][\texttt{aux}], [\texttt{question}] \\
    
    Here is the start of the article abstract for your reference: [\texttt{abstract}] 
    \end{tcolorbox}
};
\end{tikzpicture}
\caption{Standardized template used for response generation in English, analogous templates are used for other languages.}
    \label{fig:prompt}
\end{figure}

\paragraph{Prompt Design:}
This process involved carefully designing the question field for each record, which contains a natural-language query in the context of the NLP paper's scientific content.
Each question is paired with a prompt, built using a standardized template (see \Cref{fig:prompt}).
This structure ensures that every query remains explicitly grounded in the cited source, providing consistent contextual framing across languages and papers.

\paragraph{Annotation Procedure:}
Due to the large amount of generated answers, we sampled a subset of generated answers for \emph{manual annotation}. 
The sampling procedure ensures that each output is annotated only once while maintaining a balanced distribution across questions, prompts, and models. This is accomplished by randomly selecting instances and setting target annotation counts for each question–prompt–model combination, aiming for a total of eight annotations per question.

Each selected output was annotated by the same expert who created the question and evaluated along two dimensions: \emph{factual correctness} and \emph{fluency}. Factual correctness was labeled as \textit{yes} or \textit{no}, indicating whether the answer contained factual errors. Fluency was assessed on a three-point scale: \textit{yes} (well-formed), \textit{minor} (minor language issues), or \textit{no} (ungrammatical or disfluent). 

During the post annotation sanity check step, several samples containing bad generated responses were removed from en, es, hi, bn, and gu. This step resulted in final dataset distribution corresponding to \Cref{tab:dataset-stats} still keeping unique number of question as 100 for all the nine languages.

\subsection{CAP Statistics}
\label{subsec:stats}

\Cref{tab:dataset-stats} presents the statistics of the CAP dataset across the different splits and languages. In the table, \texttt{Q} denotes number of unique questions created by annotators per language and \texttt{R} denotes the number of LLM generated response annotations after post-processing step. 

\input{Tables/dataset_stats}

\Cref{tab:llms_old} presents all the different LLMs utilized in generating the response for 100 questions created per languages. All languages except French and Italian used two LLMs, whereas these used three different models. 

\input{Tables/llms_used}

\begin{figure}[ht!]
    \centering
    \includegraphics[width=\columnwidth]{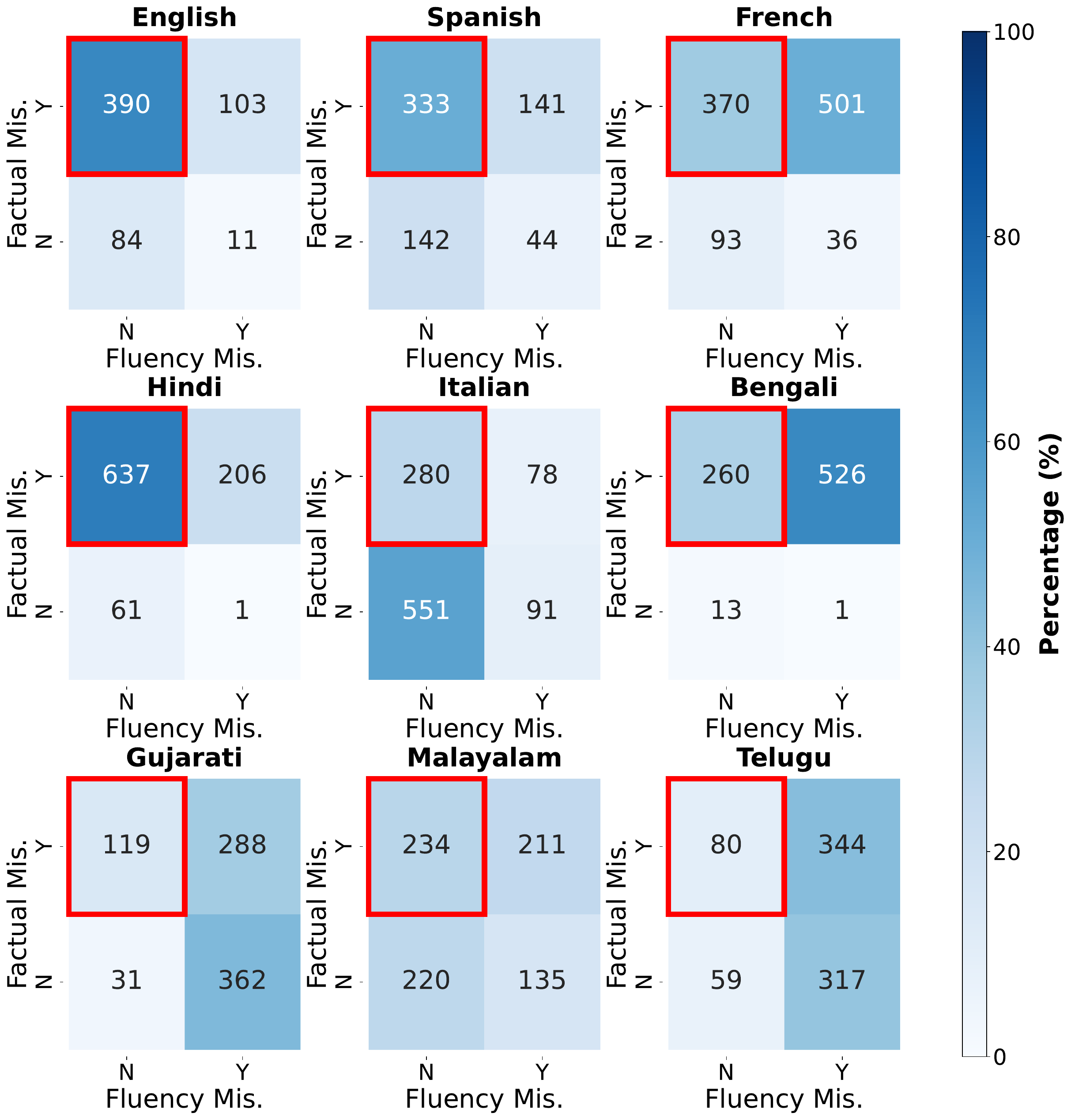}
    \caption{Label distribution per language.}
    \label{fig:flue_fact_comp}
\end{figure}

\Cref{fig:flue_fact_comp} illustrates the joint distribution of factuality and fluency annotations against each other for all the languages covered in the CAP data set. 

The formulation we adopt here is that \textit{a hallucination is defined as a response that is fluent but false} --- that is, linguistically coherent yet factually inconsistent or unsupported with respect to the underlying publication \cite{bhamidipati-etal-2024-maha}. This perspective reframes hallucination detection as a dual-facet problem, requiring models to jointly assess both the fluency and factuality of generated content.

A fact that immediately arises from observing \Cref{fig:flue_fact_comp} is that \emph{challenges vary from language to language}: for English, Spanish, French, Hindi and Bengali, models struggle to output factually correct responses; for Gujarati and Telugu, the issue is first and foremost fluency; Italian displays mostly responses that are fluent and factual, whereas Malayalam outputs are more uniformly distributed across all four possible cases. 
Hallucinations proper---i.e., outputs that are fluent but not factual---therefore have likely distinct root causes across languages.

\section{Characteristics of CAP Dataset}
\label{sec4:analysis}

CAP presents a unique and challenging dataset. To examine the factors contributing to its difficulty, we analyze citation counts (\Cref{effect-citation}) and assess how question types influence the factuality of model outputs (\Cref{subsec:taxonomy-q-type}).

\begin{figure}[ht]
    \centering
    \includegraphics[width=\columnwidth]{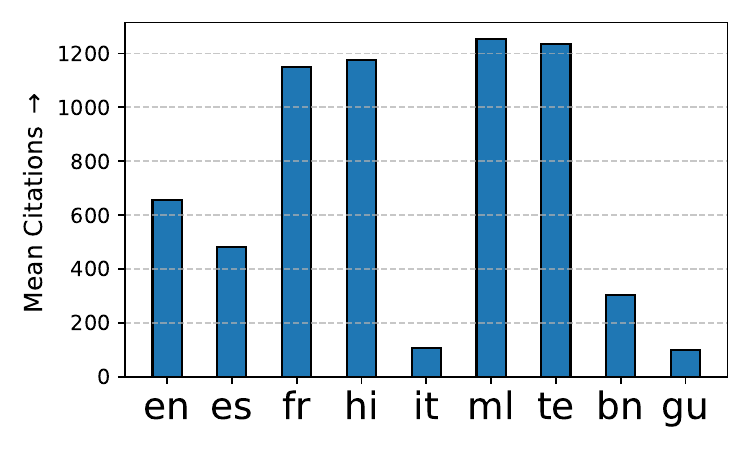}
    \caption{\label{fig:citation_distribution}Mean citation per language.}
\end{figure}

\subsection{Effects of Citation Counts}
\label{effect-citation}

An interesting side-effect of focusing on scientific publication is that we can provide rough estimates of the popularity of a certain topic, by means of bibliometric indicators.
\Cref{fig:citation_distribution} depicts the distribution of mean total citation for each language split from  the CAP data set. For example, Malayalam language data contain questions associated to NLP papers that had overall the most citations in comparison to all other languages. On the other hand, Italian and Gujarati contains questions that were created from papers which had relatively the least mean citations.

\begin{figure}[ht!]
    \centering
    \includegraphics[width=\columnwidth, trim= 0 0 0cm 0cm, clip]{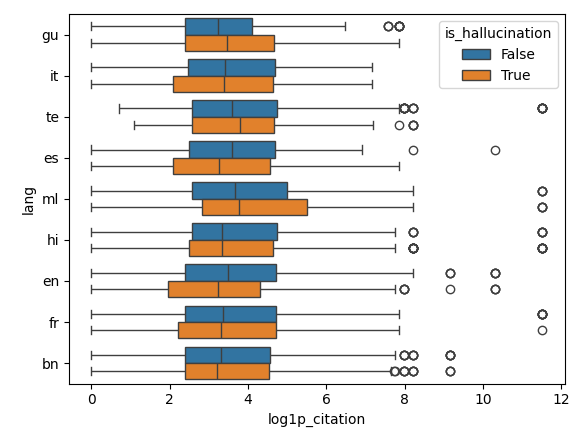}
    \caption{Distribution of citation counts per language, hallucination vs.\ non-hallucinations.}
    \label{fig:citation_versus_hallucination}
\end{figure}

Interestingly, citation counts do not seem to be indicative of hallucination. 
\Cref{fig:citation_versus_hallucination} shows the distribution of log1plus--transformed citation counts per language, according to whether the output is deemed a hallucination or not. 
As is apparent, distribution of citation counts are extremely similar for hallucinations and non-hallucinations alike.
In fact, language-specific Mann-Whitney U-tests comparing the citation counts in hallucinations vs. non-hallucinations suggest no statistically significant difference after Bonferroni correction.

\subsection{Effects of Question Types}
\label{subsec:taxonomy-q-type}

\begin{figure}[ht]
    \centering
        \includegraphics[width=\columnwidth, trim= 0 0 2cm 0.2cm, clip]{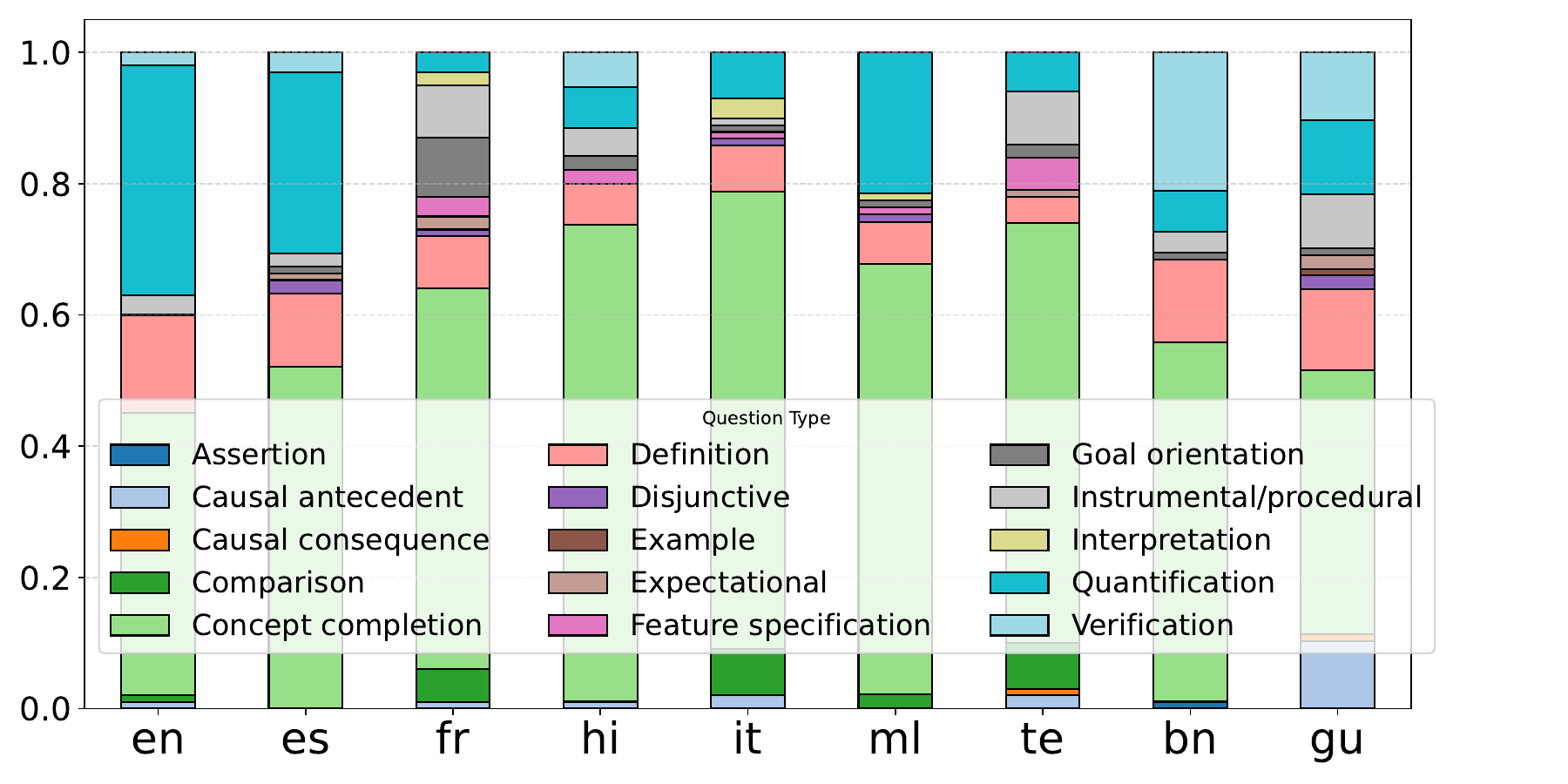}
        \caption{Question Type Distribution per language.}
    \label{fig:question_type_distribution}
\end{figure}

A major confounding factor in \Cref{fig:flue_fact_comp} is that the questions asked vary across languages, which certainly influences the outputs. To assess this point,
\Cref{fig:question_type_distribution} presents a comparison of the distribution of the different types of questions based on \citet{graesser1994question}'s taxonomy.
Specifically designed for question-answering and information retrieval tasks, this taxonomy categorizes questions into 18 distinct types, taking into account both the format of the expected answer (short or long) and its illocutionary function, such as \textit{verification}, \textit{definition}, \textit{example}, or \textit{comparison} (see also \citet{pomerantz2005linguistic}).
To automatically identify the question type for every question created by the annotators, we employed the LLM-as-a-judge approach \citep{li2024llms}, using \texttt{google/gemma-3-27b-it} \citep{team2025gemma} as the LLM judge.

Firstly, we observe that, out of the 18 classes, only 15 classes were identified across the entire dataset, with ``\textit{Concept completion}'' being the most frequent question type across all nine languages. 
The corresponding breakdown is displayed in \Cref{fig:question_type_distribution}, which also highlights the imbalance of question types in the CAP dataset. 

\begin{table}[ht!]
\centering
\small
\resizebox{\columnwidth}{!}{
    \begin{tabular}{l rrrr}
    \hline
    \rowcolor{gray!30}
    \textbf{Question Type} & \textbf{\# N} & \textbf{\# Y} & \textbf{N residual} & \textbf{Y residual} \\
    \hline
    Assertion & 0 & 1 & $-0.579$ & $0.411$ \\
    Causal antecedent & 9 & 8 & $\mathbf{1.383}$ & $-0.982$ \\
    Causal consequence & 0 & 2 & $-0.819$ & $0.581$ \\
    Comparison & 11 & 11 & $\mathbf{1.335}$ & $-0.948$ \\
    Concept completion & 146 & 360 & $-1.814$ & $1.288$ \\
    Definition & 31 & 50 & $0.738$ & $-0.524$ \\
    Disjunctive & 3 & 4 & $0.427$ & $-0.303$ \\
    Example & 1 & 0 & $\mathbf{1.148}$ & $-0.815$ \\
    Expectational & 3 & 3 & $0.697$ & $-0.495$ \\
    Feature specification & 2 & 10 & $-1.009$ & $0.716$ \\
    Goal orientation & 14 & 4 & $\mathbf{3.243}^*$ & $\mathbf{-2.303}^*$ \\
    Instru./procedural & 17 & 20 & $\mathbf{1.305}$ & $-0.927$ \\
    Interpretation & 2 & 4 & $-0.008$ & $0.006$ \\
    Quantification & 29 & 92 & $-1.816$ & $1.289$ \\
    Verification & 26 & 14 & $\mathbf{3.438}^*$ & $\mathbf{-2.442}^*$ \\
    \hline
    \end{tabular}}
\caption{Observed counts (N: no hallucination; Y: hallucination) and standardized residuals for each question type. Bold residuals indicate $|Z| \geq 1.96$ ($p<0.05$).
Per question type, $^*$ values exhibit statistically significant deviations from expected (non-)hallucination rates.
}
\label{tab:question_versus_hallucination}
\end{table}


Next, we also highlight that the type of question asked could have an effect on factuality. To examine this relationship, we perform a Chi-square test between the hallucination label (\textit{yes} or \textit{no}) and the question type, as determined through the LLM-as-a-judge evaluation, shown in \Cref{tab:question_versus_hallucination}.
The test reveals a significant association between question type and hallucination occurrence, $\chi^2$(14, N=\text{877}) = 61.21, $p<.001$.
The effect size, measured by Cramér's V, was $0.26$, indicating a small-to-medium effect according to Cohen’s guidelines.

In detail, hallucinations are not uniformly distributed across question types:
\textit{Verification} 
and \textit{Goal orientation} 
questions exhibited strong positive residuals for non-hallucinated responses and negative residuals for hallucinated responses, indicating these types are significantly less prone to hallucinations.
Conversely, \textit{Quantification} 
and \textit{Concept completion} 
showed moderate positive residuals for hallucination, suggesting a higher hallucination rate in these categories, while most other question types showed weak or no significant deviation from expected frequencies.


Together, these findings underscore the richness of the CAP dataset. By reflecting variations in citation-based content and question-type sensitivity, CAP presents a more challenging and realistic benchmark for assessing factual consistency in scientific question answering.


\section{Benchmarking on CAP}
\label{sec:analysis-model-performance}


Following \Cref{sec4:analysis}, that highlighted the complexity of CAP, 
this section evaluates the performance of existing hallucination detection tools on CAP. 

\subsection{Task Formulation}

We frame hallucination detection as a binary classification task: given a pair consisting of a scientific publication segment (premise) and an LLM-generated answer (hypothesis), the model must determine whether the hypothesis constitutes a scientific hallucination. 

\input{Tables/updated-test-benchmarks}

\subsection{Evaluation Metrics}

To ensure comparability across models and languages, we adopted a standardized processing pipeline. Each question-answer pair was segmented into smaller passages to fit model context windows, and baseline models processed each passage independently. The highest probability across passages was used as the final prediction label. We report Macro-F1 scores per language, along with the overall average for all the languages.

\subsection{Models}
We benchmark on six representative baselines spanning both \textit{reference-based} and \textit{reference-free} hallucination detection paradigms. \\

\noindent
\textbf{HHEM-2.1-Open}
The Hallucination and Hallucination Evaluation Model (HHEM-2.1-Open; hereafter referred to as HHEM) \citep{hhem-2.1-open} is an open-source model trained on factual consistency datasets to detect hallucinations in long-form outputs generated by LLMs. It functions as a strong factuality classifier, optimized using contrastive, entailment-based training objectives. \\

\noindent
\textbf{mDeBERTa-v3-base-xnli}
\citep{laurer_less_2022} is a multilingual entailment model fine-tuned on 2.7 million natural language inference (NLI) examples spanning over 100 languages. Aligned with our task formulation, the model evaluates whether a generated answer \emph{entails}, \emph{contradicts}, or is \emph{neutral} with respect to the source publication. Outputs labeled as \textit{contradiction} or \textit{neutral} are considered hallucinations.\\

\noindent
\textbf{SelfCheckGPT}
\citep{manakul2023selfcheckgptzeroresourceblackboxhallucination} is a reference-free hallucination detector that operates without access to external context. It estimates factual consistency by sampling multiple outputs from a language model and measuring the degree of internal agreement among them. We utilized a \texttt{google/gemma-2-9b} model for this, and therefore denote is as SelfCheckGemma. \\

\noindent
\textbf{XLM-RoBERTa-XL Hallucination Detector}
This multilingual reference-based hallucination detector \citep{bondarenko2024hallucination} is built on the XLM-RoBERTa-XL backbone. The model casts hallucination detection as a binary classification task, employing a self-adaptive hierarchical encoder fine-tuned in two stages: contrastive learning to optimize sentence embeddings, followed by supervised fine-tuning for classification.\\

\noindent
\textbf{FAVA} \citep{mishra2024finegrainedhallucinationdetectionediting} is a reference-based model for fine-grained hallucination detection and correction. It employs an editor language model trained on synthetically generated data to identify and revise factual inaccuracies in generated text.\\

\noindent
\textbf{HDM-2}
\citep{paudel2025hallucinothallucinationdetectioncontext} is a comprehensive hallucination detection model designed to validate outputs from LLMs using both contextual evidence and common knowledge. It employs a multi-task architecture comprising separate modules for context-based and common-knowledge verification, and generates hallucination scores at both the sentence and token levels. \\

All baselines utilize pretrained large language models in zero-shot inference setting without additional fine-tuning for fair comparison. 


\subsection{Results}
\paragraph{Disentangling the effects of fluency}
We first seek to establish the performances of existing models. In \Cref{tab:all-datapoints}, we summarize macro-F1 scores for the 6 baselines described above. 
A related point of interest is the impact of the definition of hallucination we adopt in this work, as outputs that are fluent, yet not factually correct. To disentangle the effects of fluency, in \Cref{tab:only-fluent} we also report performances on the subset of datapoints annotated as fluent, i.e., simplifying the task to one of factual correctness classification.

A few observations can be drawn from results in \Cref{tab:perfs}.
First, off-the-shelf models tend to perform remarkably poorly. Overall performances across all test datapoints are always below $0.46$. Given that we frame the problem as a binary classification, this result puts into question the efficacy of hallucination detectors in out-of-domain settings.

Second, simplifying the problem to that of a classification of factuality among fluent outputs need not yield consistent improvements. While we observe improvements for low resource Indic languages, the same does not hold consistently for high resource languages, and overall performances remain remarkably poor.
In short, the low scores we observe when measuring performance across all items are not only due to the lesser regularity of non-fluent datapoints, but rather reflects genuine difficulty intrinsic to the CAP dataset.


\input{Tables/long-context-results}

\paragraph{Effects of long-context input} 
Results in \Cref{tab:perfs} are obtained by selecting the highest probability across text chunks. While this approach is practically motivated, we can also expect it biases the models towards classifying outputs as hallucinated. 
This challenge is intrinsic to the long-context nature of the reference documents.

To assess this point more formally,
we perform an ablation study: we compare performances on a subset of items when feeding the entire article as opposed to what we would obtain by using as reference only the section deemed relevant by the annotators.
To balance annotation efforts and coverage, we focus on one high resource (English) and one low resource language (Telugu). 
We consider three reference-dependent models: Fava, HDM2 and XLMRobertaXL.

Results are listed in \Cref{tab:contxt}, presented as the margin of improvement when using as input the relevant section only. 
Following our previous approach, we report both scores on the full dataset (in \Cref{tab:contxt:all}) as well as scores obtain when considering only items presenting no issues in fluency (in \Cref{tab:contxt:fluent}).
Results suggest that performance improvements are highly contingent on the exact model considered: while Fava systematically yields lower recall and macro-F1 scores, HDM2 generally benefits from more targeted contexts.
This suggests that not all models are equally sensitive to irrelevant elements of context. 

\section{Conclusion}
We present a novel scientific hallucination dataset, the CAP dataset, comprising question–answer pairs across five high-resource and four low-resource languages. Each instance is annotated for both factuality and fluency, enabling a comprehensive evaluation of LLMs in multilingual scientific contexts. This dataset extends the scope of hallucination research to the scientific domain and promotes cross-lingual analysis of factual consistency. CAP directly addresses the growing tendency of LLMs to produce fluent yet factually incorrect content, offering a challenging benchmark for evaluating factual grounding.

Our focus on evaluating hallucination as a phenomenon at the crossroads of fluency and factuality offers an interesting complementary view to other taxonomies of hallucination. 
For instance, while the definition we rely on, in and of itself, could apply to extrinsic and intrinsic hallucinations alike (i.e., whether the hallucination directly contradicts the given input text, vs.\ whether it contradicts the LLM's training data; \citealp{ji-etal-2023-survey,bang-etal-2025-hallulens}), the dataset we provide relies mostly on knowledge obtained through exposure during pretraining or instruction-tuning (as we do not provide the entire contents of papers to the LLMs we assess) --- meaning we have focused primarily on extrinsic hallucinations.
Given this state of affairs, it is worth highlighting that our extrinsic indicator of topic popularity (viz.\ citation counts) does not appear to be a factor impacting hallucination rates, whereas intrinsic linguistic cues (such as question types) do impact LLM outputs in a measurable way.
On the other hand, the data we collect provides evidence that fluency and factuality play different roles for different languages, whereas our benchmarking experiments underscore the difficulty inherent to processing noisier, less-fluent outputs. 
Our dual-faceted outlook on hallucination therefore provides an interesting complementary lens to reassess prior research in the field.

Future work may include expanding the dataset to additional scientific disciplines such as the medical domain \citep{pal-etal-2023-med} and extremely low resource languages \citep{joshi-etal-2020-state}, exploring the interaction of hallucination with unseen languages, and leveraging CAP for developing robust, hallucination-aware generation models. 

\section{Limitations and Ethical Considerations}

Some limitations of our study concern both terminology and data scope. First, the use of the term \textit{hallucination} to describe AI-generated factual errors is inherently metaphorical and may be misleading, as it implies flawed perception rather than the statistical generation processes that underlie large language models \citep{Hicks2024}. Although we adopt this term for consistency with existing literature, we acknowledge its conceptual limitations and the potential influence such framing may have on public understanding of AI reliability. Second, the CAP dataset primarily focuses on NLP-related scientific papers, which may constrain the transferability of findings to other domains. While it spans nine languages, coverage is uneven, with lower representation and translation quality in low-resource languages. Finally, despite rigorous design and review, human annotations may still reflect subjective judgments, especially in cross-lingual assessments of factual correctness.

From an ethical perspective, the CAP dataset is derived from publicly available sources and contains no personal data. Nonetheless, model outputs may include scientifically inaccurate or misleading statements; these should not be treated as factual. The dataset is released strictly for research purposes to promote safer and more transparent scientific text generation.

\section*{Bibliographical References}
\label{sec:reference}

\bibliographystyle{lrec2026-natbib}
\bibliography{custom}


\appendix

\end{document}

%% file: Figures/sample-shroomcap-fix.tex
\begin{figure}[ht!]
    \centering
\includegraphics[page=3, 
trim = 2.55cm 15cm 10.6cm 3.7cm, clip]{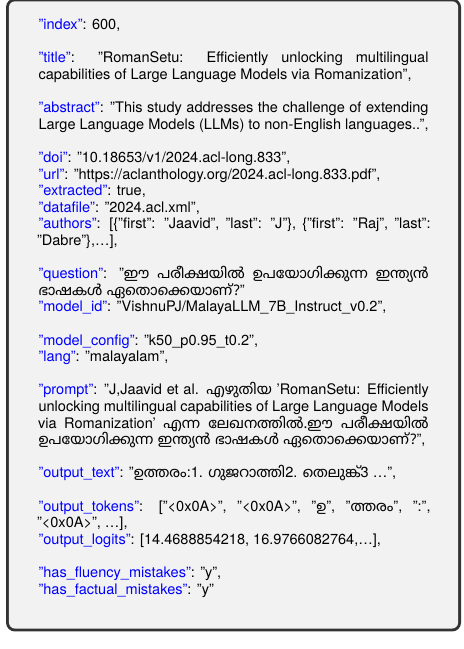}
\caption{Example instance extracted from CAP Malayalam. The question roughly translates to \textit{``Which Indian languages are used in this test set?''}. It provides all the metadata regarding the creation of this example including information related to the question and the response generated. }
\vspace{-1ex}
    \label{fig:example}
\end{figure}

%% file: Tables/dataset_stats.tex
\begin{table}[ht]
    \centering
    \resizebox{\columnwidth}{!}{
    \begin{tabular}{r cc c cc c cc c cc}
        \toprule
        \multirow{2}{*}{\textbf{lang}} &  \multicolumn{2}{c}{\textbf{TOTAL}} && \multicolumn{2}{c}{\textbf{TRAIN}} && \multicolumn{2}{c}{\textbf{VAL}} && \multicolumn{2}{c}{\textbf{TEST}}\\
        & \texttt{Q}&\texttt{R} &&\texttt{Q}&\texttt{R} &&\texttt{Q}&\texttt{R} &&\texttt{Q}&\texttt{R}  \\
        \midrule
        en & 100 & 588 && 40 & 108 && 30 & 240 && 30 & 240 \\
        es & 100 & 660 && 40 & 180 && 30 & 240 && 30 & 240 \\
        fr & 100 & 1000 && 40 & 520 && 30 & 240 && 30 & 240 \\
        hi & 100 & 905     && 40 & 425 && 30 & 240 && 30 & 240 \\
        it & 100 & 1000 && 40 & 520 && 30 & 240 && 30 & 240 \\
        \midrule
        
        **ml & 100 & 788 && ---& ---&& ---& ---&& 100 & 788\\
        **te & 100 & 800 && ---& ---&& ---& ---&& 100 & 800\\
        **bn & 100 & 798 && --- & --- && --- & --- && 100 & 798\\
        **gu & 100 & 800 && ---& ---&& ---& ---&& 100 & 800\\
        
         \bottomrule
    \end{tabular}}
    \caption{Dataset statistics per language.}
    \label{tab:dataset-stats}
\end{table}

%% file: Tables/llms_used.tex
\begin{table}[ht!]
    \centering
\resizebox{!}{.9\columnwidth}{
\begin{tabular}{l p{4.8cm} ccc }
\toprule
\rowcolor{gray!30}
         \textbf{Lang.} & \textrm{\textbf{HF identifier}} 
         & {{\textbf{N. train}}} & {{\textbf{N. val.}}} & {{\textbf{N. test}}}\\
\midrule\multirow{2}{*}{en}
& lmsys/vicuna-7b-v1.5 
&42& 121 & 120 \\
\cmidrule{2-5}
& meta-llama/Meta-Llama-3-8B-Instruct \citep{grattafiori2024llama3herdmodels} 
&66& 119 & 120 \\

\midrule\multirow{2}{*}{es}
& Iker/Llama-3-Instruct-Neurona-8b-v2 
&100& 120 & 120 \\
\cmidrule{2-5}
& meta-llama/Meta-Llama-3-8B-Instruct \citep{grattafiori2024llama3herdmodels} 
&80& 120 & 120 \\

\midrule\multirow{3}{*}{fr}
& bofenghuang/vigogne-2-13b-chat 
&174& 78 & 81\\
\cmidrule{2-5}
& occiglot/occiglot-7b-eu5-instruct 
&169& 75 & 84 \\
\cmidrule{2-5}
& meta-llama/Meta-Llama-3.1-8B-Instruct
&177& 87 & 75 \\

\midrule\multirow{2}{*}{hi}
& Cognitive-Lab/LLama3-Gaja-Hindi-8B-v0.1
&225 & 120 & 120 \\
\cmidrule{2-5}
& sarvamai/OpenHathi-7B-Hi-v0.1-Base 
& 200 & 120 & 120 \\

\midrule\multirow{3}{*}{it}
& sapienzanlp/modello-italia-9b 
&172& 86 & 81\\
\cmidrule{2-5}
& google/gemma-2-9b-it 
&183& 81 & 79 \\
\cmidrule{2-5}
& meta-llama/Meta-Llama-3.1-8B-Instruct 
&165& 73 & 80\\

\midrule\multirow{2}{*}{**ml}
& VishnuPJ/MalayaLLM-7B-Instruct-v0.2 
& --- & --- &  395\\
\cmidrule{2-5}
& sarvamai/sarvam-1 
& --- & --- &  393\\

\midrule\multirow{2}{*}{**te}
& meta-llama/Llama-3.1-8B-Instruct 
& --- & --- & 399 \\
\cmidrule{2-5}
& Telugu-LLM-Labs/Indic-gemma-7b-finetuned-sft-Navarasa-2.0 
& --- & --- & 399 \\

\midrule\multirow{2}{*}{**bn}
& BanglaLLM/BanglaLLama-3-8b-bangla-alpaca-orca-instruct-v0.0.1 
& --- & --- & 420 \\
\cmidrule{2-5}
& BanglaLLM/Bangla-s1k-qwen-2.5-3B-Instruct 
& --- & --- & 378 \\

\midrule\multirow{2}{*}{**gu}
& GenVRadmin/AryaBhatta-GemmaUltra-Merged 
& --- & --- &  600 \\
\cmidrule{2-5}
& GenVRadmin/AryaBhatta-GemmaGenZ-Vikas-Merged
& --- & --- & 200 \\
\bottomrule
    \end{tabular}
    }
    \caption{LLMs used for generating responses (\texttt{R}) for each language. 
    ** languages contain only test subset.
    }
    \label{tab:llms_old}
\end{table}

%% file: Tables/updated-test-benchmarks.tex
\begin{table*}[ht]
\centering

\begin{subtable}[t]{\textwidth}
\resizebox{\textwidth}{!}{
\begin{tabular}{
    r  
    S[table-format=2.3, round-mode=places, round-precision=3]  
    S[table-format=2.3, round-mode=places, round-precision=3]  
    S[table-format=2.3, round-mode=places, round-precision=3]  
    S[table-format=2.3, round-mode=places, round-precision=3]  
    S[table-format=2.3, round-mode=places, round-precision=3]  
    S[table-format=2.3, round-mode=places, round-precision=3]  
    S[table-format=2.3, round-mode=places, round-precision=3]  
    S[table-format=2.3, round-mode=places, round-precision=3]  
    S[table-format=2.3, round-mode=places, round-precision=3]  
    S[table-format=2.3, round-mode=places, round-precision=3]  
}
\toprule
\rowcolor{gray!30}
\textbf{Model} & {\textbf{en}} & {\textbf{es}} & {\textbf{fr}} & {\textbf{hi}} & {\textbf{it}} & {\textbf{ml}} & {\textbf{te}} & {\textbf{bn}} & {\textbf{gu}} & {\textbf{Overall}} \\
\midrule
HHEM & 0.4857 & 0.4198 & 0.3571 & 0.4455 & 0.2584 & 0.3662 & 0.0933 & 0.2472 & 0.2487 & 0.2939 \\
mDeBERTaNLI & 0.4346 & 0.4972 & 0.5857 & 0.4626 & 0.5345 & 0.3158 & 0.2143 & 0.4995 & 0.2536 & 0.3891 \\
SelfCheckGemma & 0.503 & 0.4939 & 0.5201 & 0.556 & 0.3941 & 0.4544 & 0.4086 & 0.4937 & 0.3283 & 0.4588 \\
XLMRobertaXL& 0.4401 & 0.4165 & 0.2427 & 0.4045 & 0.2035 & 0.3899 & 0.136 & 0.4067 & 0.2775 & 0.3336 \\
FAVA & 0.4329 & 0.5251 & 0.3119 & 0.4896 & 0.2714 & 0.4302 & 0.3796 & 0.5081 & 0.3571 & 0.4558 \\
HDM2 & 0.4592 & 0.4246 & 0.2773 & 0.4434 & 0.2764 & 0.3954 & 0.2976 & 0.5235 & 0.2599 & 0.3979 \\
\bottomrule
\end{tabular}}
\caption{\label{tab:all-datapoints}Macro-F1 metric, considering both classes.}
\end{subtable}

\begin{subtable}[t]{\textwidth}
\resizebox{\textwidth}{!}{
\begin{tabular}{
    r  
    S[table-format=2.3, round-mode=places, round-precision=3]  
    S[table-format=2.3, round-mode=places, round-precision=3]  
    S[table-format=2.3, round-mode=places, round-precision=3]  
    S[table-format=2.3, round-mode=places, round-precision=3]  
    S[table-format=2.3, round-mode=places, round-precision=3]  
    S[table-format=2.3, round-mode=places, round-precision=3]  
    S[table-format=2.3, round-mode=places, round-precision=3]  
    S[table-format=2.3, round-mode=places, round-precision=3]  
    S[table-format=2.3, round-mode=places, round-precision=3]  
    S[table-format=2.3, round-mode=places, round-precision=3]  
}
\toprule
\rowcolor{gray!30}
\textbf{Model} & {\textbf{en}} & {\textbf{es}} & {\textbf{fr}} & {\textbf{hi}} & {\textbf{it}} & {\textbf{ml}} & {\textbf{te}} & {\textbf{bn}} & {\textbf{gu}} & {\textbf{Overall}} \\
\midrule
HHEM & 0.467 & 0.4646 & 0.4306 & 0.5535 & 0.3022 & 0.5434 & 0.4114 & 0.5294 & 0.4000 & 0.5141 \\
mDeBERTaNLI & 0.4215 & 0.4577 & 0.5415 & 0.394 & 0.552 & 0.4016 & 0.3965 & 0.4723 & 0.5122 & 0.5184 \\
SelfCheckGemma & 0.518 & 0.4572 & 0.5888 & 0.5188 & 0.384 & 0.5185 & 0.5773 & 0.4977 & 0.4802 & 0.5296 \\
XLMRobertaXL & 0.4752 & 0.4237 & 0.4132 & 0.4684 & 0.2252 & 0.3568 & 0.3594 & 0.4862 & 0.4519 & 0.4072 \\
FAVA & 0.453 & 0.5383 & 0.4681 & 0.671 & 0.3052 & 0.4107 & 0.5087 & 0.4613 & 0.4648 & 0.4752 \\
HDM2 & 0.4928 & 0.436 & 0.4261 & 0.4575 & 0.3165 & 0.4188 & 0.3928 & 0.5478 & 0.5032 & 0.4595 \\

\bottomrule
\end{tabular}}
\caption{\label{tab:only-fluent}Macro-F1 metric, considering only fluent datapoints.}
\end{subtable}

\caption{Performance comparison between hallucination detectors across languages.}
\label{tab:perfs}
\end{table*}

%% file: Tables/long-context-results.tex
\begin{table}[ht!]
    \centering

\begin{subtable}[t]{\columnwidth}
\centering
\begin{tabular}{lr *{3}{S[table-format=2.3, round-mode=places, round-precision=3]}}
\toprule
\rowcolor{gray!30}
 & \textbf{Model}  & {{\textbf{F1}}} & {{\textbf{Rec}}} & {{\textbf{Prec}}} \\
\midrule
\multirow{3}{*}{\textbf{en}} & FAVA & -0.025882 & -0.056478 & -0.022812 \\
 & HDM2 & 0.118349 & 0.060631 & 0.015787 \\
 & XLMRobertaXL & 0.018856 & 0.008929 & 0.501895 \\
\midrule
\multirow{3}{*}{\textbf{te}} & FAVA & 0.156818 & 0.059308 & 0.034877 \\
 & HDM2 & 0.115097 & 0.034022 & 0.001443 \\
 & XLMRobertaXL & 0.005704 & 0.002688 & 0.000252 \\
\bottomrule
\end{tabular}
    \caption{All datapoints}
    \label{tab:contxt:all}
\end{subtable}

\begin{subtable}[t]{\columnwidth}
\centering
\begin{tabular}{lr *{3}{S[table-format=2.3, round-mode=places, round-precision=3]}}

\toprule
\rowcolor{gray!30}
 & \textbf{Model}  & {{\textbf{F1}}} & {{\textbf{Rec}}} & {{\textbf{Prec}}} \\
\midrule
\multirow{3}{*}{\textbf{en}} & Fava & -0.047005 & -0.019211 & -0.004542 \\
 & HDM2 & 0.160869 & 0.170880 & 0.049651 \\
 & XLMRobertaXL & 0.043306 & 0.021739 & 0.502810 \\
\midrule
\multirow{3}{*}{\textbf{te}} & Fava & -0.095891 & -0.013542 & 0.035207 \\
 & HDM2 & 0.088035 & 0.020312 & 0.068190 \\
 & XLMRobertaXL & 0.000000 & 0.000000 & 0.000000 \\
\bottomrule
\end{tabular}
\caption{Only fluent datapoints}
\label{tab:contxt:fluent}
\end{subtable}
\caption{Effects of narrowing the context (scores when using the relevant context as input minus scores when using the full paper).}
\label{tab:contxt}
\end{table}